# Leveraging Large Language Models for Robot-Assisted Learning of Morphological Structures in Preschool Children with Language Vulnerabilities

Stina Sundstedt [1(✉) [0000-0002-3895-0760]], Mattias Wingren [2[0000-0003-4269-5093]], Susanne Hägglund [2[0000-0002-4941-6811]], and Daniel Ventus [2[0000-0002-2347-8213]]

[1] Faculty of Arts, Psychology and Theology, Åbo Akademi University, 20500 Turku, Finland
[2] Experience Lab, Faculty of Education and Welfare studies, Åbo Akademi University, Akademi University, 65100 Vaasa, Finland
`{stina.sundstedt, mattias.wingren, susanne.hagglund, daniel.ventus}@abo.fi`

**Abstract.** Preschool children with language vulnerabilities—such as developmental language disorders or immigration related language challenges—often require support to strengthen their expressive language skills. Based on the principle of implicit learning, speech-language therapists (SLTs) typically embed target morphological structures (e.g., third person -s) into everyday interactions or game-based learning activities. Educators are recommended by SLTs to do the same. This approach demands precise linguistic knowledge and real-time production of various morphological forms (e.g., "*Daddy wears these when he drives to work*"). The task becomes even more demanding when educators or parent also must keep children engaged and manage turn-taking in a game-based activity. In the TalBot project our multiprofessional team have developed an application in which the Furhat conversational robot plays the word retrieval game "Alias" with children to improve language skills. Our application currently employs a large language model (LLM) to manage gameplay, dialogue, affective responses, and turn-taking. Our next step is to further leverage the capacity of LLMs so the robot can generate and deliver specific morphological targets during the game. We hypothesize that a robot could outperform humans at this task. Novel aspects of this approach are that the robot could ultimately serve as a model and tutor for both children and professionals and that using LLM capabilities in this context would support basic communication needs for children with language vulnerabilities. Our long-term goal is to create a robust LLM-based Robot-Assisted Language Learning intervention capable of teaching a variety of morphological structures across different languages.

**Keywords:** human-robot interaction, language vulnerability, large language model, robot assisted language learning, social robot, speech language pathology intervention,

## 1      Introduction

Children with language vulnerabilities, such as developmental language disorder (DLD), neuropsychiatric conditions, or challenges stemming from migration, often require tailored support to foster their language development. The mechanisms behind the vulnerabilities are different depending on the population. Children with DLD have impaired abilities to obtain morphological and syntactic skills [1–3], children with autism may have very varying language skills and an atypical path of language acquisition [4], and some children with a migrant background, who are learning the school language as a second language outside the home or are newly arrived, may require additional support and tailored instruction to acquire an age-appropriate language skill in the second language.

Speech-language therapists (SLTs) play a pivotal role in delivering interventions, collaborating closely with caregivers and educators. To reinforce language learning, SLTs often encourage educators to provide examples of specific linguistic structures, such as morphological structures that the child needs to practice, during everyday interactions. However, this is challenging, as it requires not only linguistic knowledge but also the ability to produce consistent and diverse examples in real time.

## 2      The Challenge with Existing Interventions

Language interventions span an implicit-explicit continuum, with implicit approaches fostering unconscious knowledge acquisition through exposure, and explicit methods involving deliberate instruction [5]. The optimal balance of implicit and explicit strategies for teaching children morphology remains unclear [6, 7]. Research on children with developmental language disorders suggests that implicit interventions using a greater variety of words with the target morphology are more effective than those with limited variability [8].

Game-based learning is widely used in speech and language therapy to engage children [9]. For instance, in Finland, a word retrieval game named Alias is commonly used where an SLT provides descriptive clues, and the child tries to guess the corresponding word, e.g. "*This is a pet with soft fur that likes to eat carrots*". This game can be used to target semantic hierarchies like pets, vehicles, vegetables. Additionally, it can be enhanced with implicit techniques to target morphological structures within the game. For instance, an SLT might give the following clue targeting the third person singular -s morphological structure:" *This big red vehicle drive**s** through the city when there's an emergency. When it arrive**s** at a fire, it spray**s** water to put out the fire".* While playing, it is also possible to incorporate explicit instructions, enhancing the learning experience further. E.g. "*When we talk about something or someone doing an action we add -s at the end of the word.*".

While seemingly simple games like these can incorporate multiple intervention targets, implementing them effectively is very demanding. SLTs are trained to model morphological rules in therapy, yet it is challenging to spontaneously generate sentences with diverse target words, especially while providing a high dose of specific morphological structures. Imagine being asked to provide clues for the word bedroom while simultaneously being expected to use several different irregular plural nouns in the clues. If this also needs to occur in a game-based context, where multiple children are involved, and the adult must manage turn-taking and maintain the children's motivation, the task becomes significantly demanding even for experienced professionals.

The challenge grows when SLTs collaborate with caregivers and educators to support children outside therapy sessions. Educators are often expected to reinforce exercises provided by SLTs [10], but adherence can falter due to insufficient resources, guidance, or competence. Moreover, some children may not receive direct SLT support due to limited availability. These barriers highlight the need for scalable alternatives to support language intervention delivery.

## 3     Robot Assisted Language Learning as an Alternative to Support Language Intervention

Robot-Assisted Language Learning (RALL) has emerged as a promising avenue to address some of the challenges in delivering effective language interventions. Social or conversational robots have been employed in the roles of tutors, peers or teacher assistants in second-language learning [11, 12] and as tools for language and social interaction interventions for children with developmental language disorders or autism [13, 14]. While studies have shown that some SLTs express reservations about giving up control of the therapeutic situation, they acknowledge the significant potential of social robots to benefit children [14].

Many previous studies have relied on static interactions, such as video dramas featuring robots, or teleoperated robots controlled by humans [12, 13, 15–17]. While these approaches allow for controlled and predictable interactions, they lack the adaptability and responsiveness that is needed when interacting with children with language vulnerabilities. Fully autonomous systems, while offering greater independence, have traditionally been constrained by rigid, rule-based programming making them less flexible in managing spontaneous interactions and possible errors. Various challenges have been identified in previous research on RALL. Key issues highlighted in research include insufficient speech recognition [18, 19] as well as the generation and timing of verbal and non-verbal actions [11]. For example, managing turn-taking in a multi-party game is a particularly complex non-verbal task for a conversational robot [20]. These challenges are likely to be even more pronounced when working with children who have language vulnerabilities, as their specific needs may exacerbate existing limitations in robot communication.

Given the existing challenges with RALL, our team saw the possibilities of using LLMs to enhance RALL for children with language vulnerabilities. We believe that incorporating LLMs into conversational robots can help address several of the difficulties previously mentioned, making RALL more effective and accessible for this group.

## 4    Using an LLM in a RALL application - TalBot

### 4.1    Background and Ethical Considerations

In 2024, our multiprofessional research group, consisting of an SLT, a psychologist, a cognitive scientist, and a sociologist, began developing TalBot, a robot application for language intervention that integrates use of a large language model (LLM, ChatGPT) with the Furhat conversational robot [21]. Our first prototype has been iteratively tested and co-designed together with 13 preschool children (5.5-6.5 years old) with language vulnerabilities and 7 educators (4 educators and 3 educator students), who were interviewed regarding their experiences and views on the robot's performance.

The study was conducted with ethical approval from the Board for Research Ethics at Åbo Akademi University. Before the project began, a risk assessment was carried out to evaluate the potential data protection risks of using LLMs with a vulnerable group. To protect the children's personal data, the task was designed in such a way that they did not need to share any personal information when interacting with the robot. If a child began discussing topics unrelated to the game, the LLM redirected the conversation back to the game. Test facilitators were prepared to intervene if a child was at risk of continuing to share personal information. The only personal data temporarily processed by a third party was the children's voice data, which was handled by Microsoft Azure's voice recognition services.

### 4.2    TalBot in Action

```
Robot via text to speech: "Here is a new word! This is an animal
living in foreign countries. It has stripes and is often orange
and black. Which animal could it be?"

Child: "tigel"

Speech recognition to LLM: "dig"*

LLM generated text: "[excited] Almost correct! I think you mean
"tiger". Good try! Tigers are big cats with stripes. Next player,
your turn [next_turn]"

Robot via text to speech: "Almost correct! I think you mean
"tiger". Good try! Tigers are big cats with stripes. <Robot shows
exited facial gesture because of the [excited]-emotion label.>
"Next player, your turn!" <Robot indicate change of turn by turn-
ing gaze and head towards the next child as a result of the
[next_turn] command>
```

**Fig. 1.** Excerpt from our test with preschool children with language vulnerabilities. Translated from Swedish. Turn-taking and emotion labels are green and marked with [ ]. Descriptions of robot behavior are marked with grey color and < >. *The voice recognition fails to correctly recognize the correct word due to poor pronunciation.

In our novel setup, we combine the Furhat robot's conversational- and interaction-related capabilities with the generative power of an LLM. This enables TalBot to lead the Alias game, manage turn-taking, provide feedback, offer additional clues when needed, while enhancing the interaction by providing appropriate facial expressions. During the activity, TalBot plays the language retrieval game Alias with three children, providing clues while they take turns guessing the correct word (See fig 1 for an example of a whole turn of Alias).

The interplay between Furhat's capabilities and the LLM capabilities is summarized in Figure 2. Our tests with the target group of children with language vulnerabilities indicate that the TalBot application can manage turn-taking with three preschool children, and the application shows promising potential in compensating for children's language difficulties and speech recognition limitations. It can also interpret mispronunciations or mistakes by the voice recognition and adjust its responses accordingly.

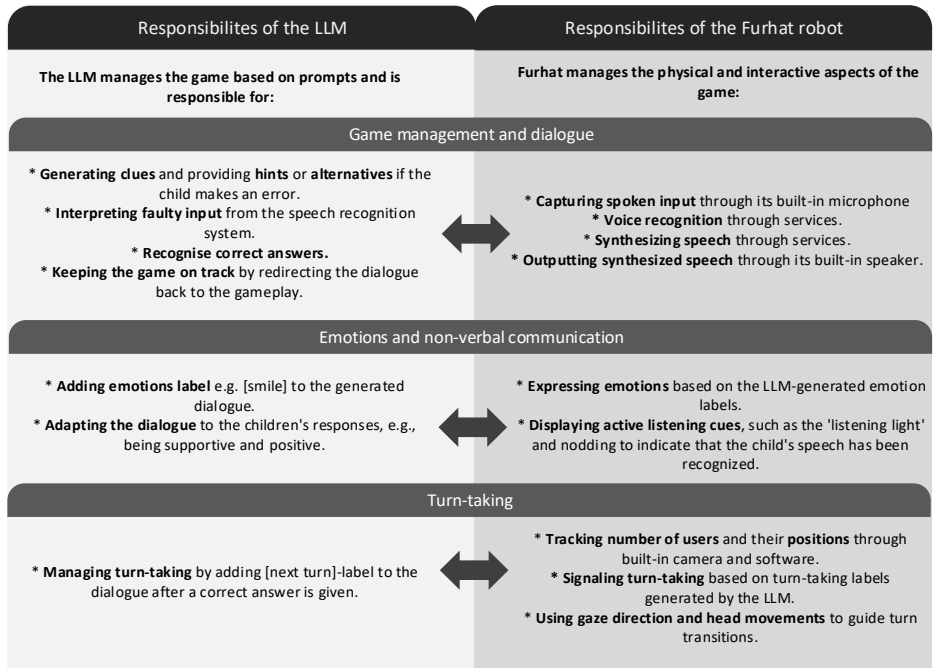

**Fig. 2.** Illustration of how the TalBot robot application uses an interplay between the capabilities and responsibilities of the LLM and the Furhat robot.

### 4.3    Advancing RALL with LLMs

Our work with TalBot advances the current state-of-the-art by addressing several key limitations in earlier RALL approaches. As part of the first wave of LLM-powered robots, TalBot enhances adaptability and fosters more natural interactions in learning contexts. Traditional RALL systems are largely rule-based, meaning they can only process speech effectively if it follows strict predefined patterns [12]. Some systems incorporate Intent Recognition, but these too rely on pre-set intents and responses, and the endless possibilities of real world utterances are still challenging to manage [22].

LLMs, on the other hand, can handle a wide range of inputs without depending on predefined structures. This adaptability is particularly important for our target group, whose speech is often unclear, a difficulty further exacerbated by the limitations of speech recognition technology.

The ability to handle turn-taking in the Alias game is important to ensure that the children taking part in the game gets equal opportunities to participate in the guessing. Managing turn-taking in an Alias game would be highly challenging without an LLM, as traditional systems require predefined target responses e.g., the correct answer in the game. The advantage of our approach is that responses do not need to be exact matches—only close approximations. With contextual cues, the LLM can often infer the intended word, even if the child mispronounces it or if voice recognition falls short (see Figure 1 for an example).

Another strength of our approach is that the rules of Alias clearly dictate turn-taking, making it easier for the robot to manage. The LLM only needs to determine whether the correct word has been said, without having to navigate the more complex turn-taking rules of spontaneous conversation [20].

Additionally, LLMs offer a more natural and adaptive speech experience compared to earlier approaches, as responses are not predefined. This is particularly beneficial in our case, as the LLM can handle unrelated responses and seamlessly steer the conversation back to the task—something that traditional RALL systems would likely struggle with.

Finally, instead of manually programming the rules for how the Alias game should proceed, an LLM can lead the game more dynamically by following a prompt that describes the task. This makes gameplay more fluid and interactive while also simplifying the process of adding new words and categories—tasks that would have been time-consuming with previous methods. By leveraging the capabilities of LLMs to control the robot's dialogue, behavior, and game content, this approach unlocks new possibilities for language-learning interventions.

Based on our experiences with LLM-based RALL, morphology learning appears to be a particularly promising areas for further exploration, especially given the complexity of providing consistent and targeted linguistic input in interactive settings (e.g., third person -s). By leveraging LLMs, conversational robots could help address some of the inherent challenges in producing and modeling morphological structures in real time. Uniquely, the robot could act as a linguistic model for children by generating and delivering structured and repetitive exposure to key grammatical forms within a game-based framework. We hypothesize that an LLM-based robot could outperform a human in this role and thus possibly serve as a role model for the professional alleviating some of the cognitive demands tied to real-time language modeling.

## 5     Exploration and Possible Implementation of Using an LLM-based RALL for Morphological Learning

Our next step for TalBot is to expand its scope to include grammar and morphological learning, positioning it as a particularly promising tool for clinical interventions. This novel approach would support the morphological language learning in three ways: generation of linguistic material, providing a linguistic model and acting as a role model for educators and other professionals.

### 5.1     Generation of linguistic material

Firstly, an LLM can produce unlimited amounts of grammatically correct linguistic output that can be used as linguistic model in gameplay. Before the advent of generative approaches, this required manually creating language material and programming distinct rule-based interactions - an approach that was both time-consuming for SLTs and developers and inherently rigid. With this older method, a preschool implementing a RALL system would eventually exhaust its pre-programmed material, potentially hindering both motivation and learning. However, with LLMs, this challenge is significantly reduced. Material can be created faster and be adapted to different special interests and needs. For example, an LLM can generate customized clues for language-learning games, such as the following clue for the word trousers: "*Daddy Pig wears these to work. Peppa sees that George likes shorts, but Daddy Pig always chooses these. They cover both legs and keep him warm. What are they*?"

### 5.2     Providing a Linguistic Model

Secondly, for a human it is both time consuming and challenging to provide a consistent linguistic model with a high dose of the targeted morphology. To be effective, one must learn the linguistic material and deliver it tirelessly. While this may seem straightforward to the uninitiated, it is difficult to consistently produce sentences following a specific pattern—for example, using only one morphological structure (e.g., past tense verbs or plural nouns).

For a robot, however, this task is trivial. Unlike humans, it does not struggle with speaking in an unnatural way or maintaining strict linguistic patterns. In this context, a robot's rigidity becomes a strength. Additionally, a robot never tires, whereas morphological learning requires extensive exposure to varied structured input [8], demanding more time than preschool educators and speech-language therapists typically have available.

### 5.3     Acting as a Role Model for Educators and other Professionals

Thirdly, given the challenges of delivering high-quality morphological interventions in game-based settings, an LLM-based robot could serve as a valuable role model for educators and other professionals by demonstrating structured linguistic input. Educators could benefit from observing the robot as it delivers consistent and targeted language models, allowing them to adopt similar sentence structures and strategies in other classroom interactions. Language learning is an ongoing process that extends beyond designated SLT intervention hours—it occurs throughout the entire school day. Therefore, relying solely on occasional therapy sessions is not sufficient.

By integrating a robot into the learning environment, preschool educators could receive continuous modeling. Instead of requiring an SLT to visit preschools to demonstrate interventions, the robot could at least partly take on this role, reducing the need for external training while maintaining high-quality instruction. This scalable approach could help bridge the gap in language support, ensuring that children receive more frequent and structured exposure to morphological patterns in naturalistic learning environments.

### 5.4    Implementation and Evaluation

Going forward, we will continue to develop the TalBot RALL-intervention through a human-centered approach, organized into four interconnected phases. Initially, we will continue our laboratory studies with children and educators to ensure that the language intervention works as intended and to identify areas for improvement in its interaction capabilities. During this phase, we will apply insights from existing literature on morphological interventions while exploring the ability of LLMs to generate linguistic models for morphology. Based on this foundation, we will identify and test appropriate morphological targets for the intervention, ensuring that the approach is both theoretically grounded and practically applicable.

Following the lab work, we will conduct field studies in preschools, engaging children with various language vulnerabilities whose primary school language is Swedish. During these studies, children will participate in structured 15-minute sessions, interacting either with our TalBot application or with an educator. We will compare how effectively the robot and educators produce specific morphological structures (e.g., verb forms or pluralization) within the context of a word retrieval game. We analyze the within session dose of the target morphological structures (e.g., number of specific verb or plural forms per unit of time) by counting their occurrences in the robot's text logs and transcripts from the educator-led sessions, providing an objective and quantifiable metric to evaluate effectiveness in consistently modeling and delivering targeted language input.

During these activities, we will collect video recordings from cameras positioned at the front and rear, audio recordings of interviews, and detailed interaction logs generated by the robot. These data sources will help us systematically analyze and quantify how frequently the robot and the educators successfully models targeted morphological forms. We will also analyze if observing the robot as a role model will affect the educators' performance. Video material and interviews are analyzed by several researchers following principles of conversation and content analysis.

The field study is crucial, as the robot needs to be tested in the environment where it is intended to be used. Additionally, we will examine what training and support educators need to effectively use the robot. This part of our study will provide insights into practical challenges and aspects that need further development to ensure the robot's successful integration into the school environment. Insights gained during the field study will inform another round of laboratory-based co-development of the intervention together with children and educators.

In the subsequent phase, we will perform a feasibility study. Here, we aim to recruit 5 to 10 children with clinically significant language vulnerabilities. We will assess their morphological skills both before and after the intervention, using established tests. The sessions will again be recorded and analyzed, with a particular focus on observing changes in interaction over time. Additionally, we will interview educators who work alongside the robot to gain insights into their experiences and perspectives on collaboration.

Throughout all phases, we will continually assess the perceptions of users—children and educators—regarding the robot's capabilities in gameplay, speech, and listening, as well as their emotional responses such as enjoyment or fear. Upon completion of these activities, we will have thoroughly developed and evaluated the intervention, preparing it for potential future studies, including larger randomized controlled trials. Our long-term goal is to create a robust LLM-based RALL intervention capable of teaching a variety of morphological structures across different languages.

### 5.5    Considerations

**Regulations**. A more personalized learning would require sharing user data with the LLM, which is problematic regarding data protection. A possible solution is running the LLM locally; however, current Swedish language models are not yet sufficiently advanced. Our current approach is therefore to minimize the data shared with the LLM-services. The risk-based rules set out in the EU AI Act must also be considered in the design of the Talbot-application.

**Ethical Considerations.** As Belpaeme et al. [11] highlight, long-term effects of robot-assisted learning on child development must be assessed. During our project we will analyze the impact of AI on children through UNICEF canvas for operationalization of policy guidelines for AI and children, and ALTAI ethical framework.

**Practical implementation in educational settings.** There is also the possibility of resistance, both in the sense that educational settings may be reluctant to use the intervention, and that it may be challenging to motivate the children to participate actively during a long-term intervention. The high cost of conversational robots may also hinder their integration into educational settings. However, insights from this project can also inform the development of alternative interventions using smartphones, computer-based games, or traditional speech therapy.

## 6    Conclusion

Building on our laboratory testing of the TalBot robot application, we plan to explore the development of an intervention targeting morphological learning for children with language vulnerabilities. This paper outlines a novel approach to integrating LLMs into RALL interventions, aiming to address key challenges in providing structured linguistic modeling in this target group. By leveraging the capabilities of LLMs, we propose a scalable and adaptive application that can support both implicit and explicit language learning while serving as a role model for educators. We believe that integrating an LLM in this context could be highly beneficial, as its performance may exceed that of a human in generating and delivering structured linguistic models for morphology. A key aspect of this work is continuous collaboration with children and educators in their everyday environments through a co-creation process, while also carefully considering ethical risks and data protection for this vulnerable group.

While this approach holds significant promise, several challenges remain, including ethical considerations, data security, and practical implementation in educational settings. Our plan is to develop and evaluate the intervention through both lab and field studies, with the long-term goal of creating a robust LLM-based RALL application capable of teaching a variety of morphological structures across different languages.

**Acknowledgments.** This study was funded by Högskolestiftelsen i Österbotten. We thank the City of Vaasa, the children, and the educators for their contributions to TalBot's development

**Disclosure of Interests.** The authors have no competing interests to declare that are relevant to the content of this article.